\documentclass[journal]{IEEEtran}
\usepackage[numbers,sort&compress]{natbib}
\usepackage{graphicx}
\usepackage{algorithmic}
\usepackage{algorithm}
\usepackage{multirow}
\usepackage{multicol}
\ifCLASSINFOpdf
\else
\fi
\usepackage{amsmath}
\begin{document}
%
\title{Rethink Maximum Mean Discrepancy for Domain Adaptation}
%
%
%
\author{Wei~Wang,
		Haojie~Li$^\ast$,
        Zhengming~Ding,
        and Zhihui~Wang
\thanks{W. Wang, H. Li and Z. Wang were with the DUT-RU International School of Information Science \& Engineering, Dalian University of Technology, Dalian, Liaoning, 116000, P.R. China e-mail: (WWLoveTransfer@mail.dlut.edu.cn, hjli@dlut.edu.cn, zhwang@dlut.edu.cn).}
\thanks{Z. Ding was with the Department of Computer, Information and Technology, Purdue School of Engineering and Technology, Indiana University-Purdue University Indianapolis, Indianapolis, IN, 46202, USA e-mail: zd2@iu.edu.cn}}

%
%

\markboth{IEEE TRANSACTIONS ON ***}%
{Wang \MakeLowercase{\textit{et al.}}: Bare Demo of IEEEtran.cls for IEEE Journals}
%



\maketitle

\begin{abstract}
Existing domain adaptation methods aim to reduce the distributional difference between the source and target domains and respect their specific discriminative information, by establishing the Maximum Mean Discrepancy (MMD) and the discriminative distances. However, they usually accumulate to consider those statistics and deal with their relationships by estimating parameters blindly. This paper theoretically proves two essential facts: 1) minimizing the MMD equals to maximize the source and target intra-class distances respectively but jointly minimize their variance with some implicit weights, so that the feature discriminability degrades; 2) the relationship between the intra-class and inter-class distances is as one falls, another rises. Based on this, we propose a novel discriminative MMD. On one hand, we consider the intra-class and inter-class distances alone to remove a redundant parameter, and the revealed weights provide their approximate optimal ranges. On the other hand, we design two different strategies to boost the feature discriminability: 1) we directly impose a trade-off parameter on the implicit intra-class distance in MMD to regulate its change; 2) we impose the similar weights revealed in MMD on inter-class distance and maximize it, then a balanced factor could be introduced to quantitatively leverage the relative importance between the feature transferability and its discriminability. The experiments on several benchmark datasets not only prove the validity of theoretical results but also demonstrate that our approach could perform better than the comparative state-of-art methods substantially.
\end{abstract}


%
\IEEEpeerreviewmaketitle

\section{Introduction}
%
%
%
%
\IEEEPARstart{D}ue to the available substantial amount of labeled data, traditional machine learning algorithms have achieved remarkable performances on the object recognition task when the training and test data follow the same or similar distributions. In most realistic scenarios, though, only a fully-labeled source domain is available to us, from which we do wish to learn a transferable classifier from the source domain, to correctly predict the data labels for a new target domain with different distribution \cite{ACM1, ACM2, ACM3}. Fortunately, Domain Adaptation (DA) as a novel emerging technique has prepared us for this pressing challenge, or rather, it has been committed to narrowing the distributional differences between the source and target domains so that extensive knowledge in the source domain could be desirably transferred to the target, thus the relabeling consumption is mitigated largely \cite{ACM4, ACM5, ACM6}.

\begin{figure}[!t]
	\centering
	\includegraphics[width=1.0\linewidth,height=0.16\textheight]{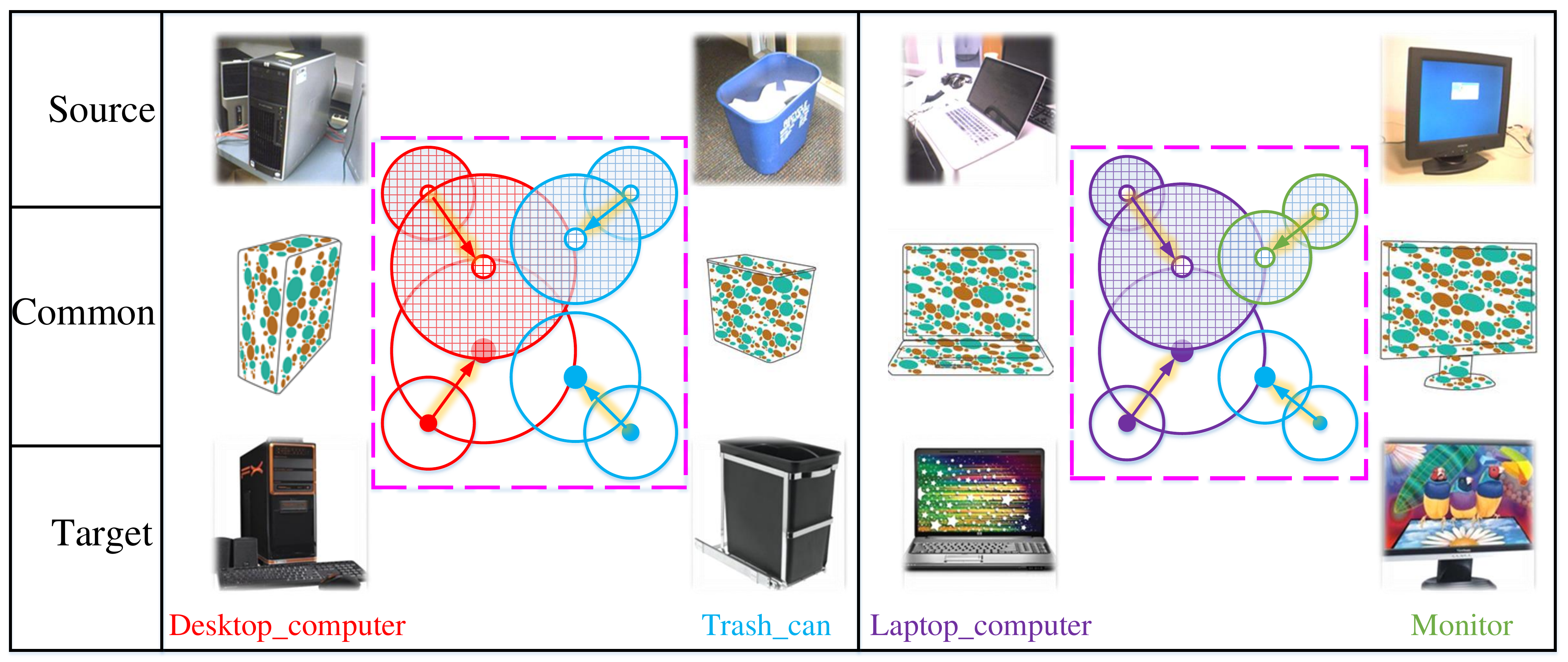}
	\caption{Working principle of the MMD. Different color circles represent various categories, the tiny circles are the means of specific categories, the hollow and meshed circles represent the source and target domains, the solid arrows denote the DA processes with MMD, and the comparatively larger circles are the transformed data features.}
	\label{fig1}
\end{figure}

\par An essential issue in the DA is to formulate an appropriate metric of the distributional distance to be used for measuring the proximity of two different distributions. Over the years, numerous distance-metric methods were proposed. For example, the Quadratic \cite{BD}, Kullback-Leibler \cite{KL}, Mahalanobis \cite{RTML} distances that are derived from the Bregman Divergence but generated by different convex functions were introduced to explicitly match the two different distributions. However, it is inflexible to extend them into different DA models due to their inconvenient manipulation. Additionally, on account of theoretical deficiencies, they are unable to describe more complicated distributions such as conditional and joint distributions. The Wasserstein distance from optimal transport problem exploited a transportation plan also aligning the two different distributions \cite{WD}, but it is still very tedious to be applied to subspace-learning DA methods because it will come down to a complex bi-level optimization problem that is nontrivial to optimize.

\par Noticeably, the Maximum Mean Discrepancy (MMD) \cite{MMD}, a metric based on the embedding of distribution measures in a reproducing kernel Hilbert space, has been applied successfully in a wide range of problems thanks to its simplicity and solid theoretical foundation, such as transfer learning \cite{TCA}, kernel Bayesian inference \cite{KBP}, approximate Bayesian computation \cite{ABC}, two-sample \cite{MMD}, goodness-of-fit testing \cite{GFT}, MMD GANs \cite{GMMN}, and auto-encoders \cite{IFVAE}, etc. For the DA setting, MMD aimed to minimize the deviation of means that are respectively computed by the source and target domains. Together with label constraint, the class-wise MMD \cite{JDA} further diminished the mean deviations of each two coupled classes with the same labels but respectively from the two different domains. Moreover, MMD and class-wise MMD are usually adopted to respectively measure the marginal and conditional distribution differences between the two different domains \cite{JDA}. Notably, this paper only concentrates on rethinking the class-wise MMD since MMD is a special case of the class-wise MMD when we regard a domain as one category, and the DA performances are mainly determined by the class-wise MMD.   

\par The early DA methods usually directly integrated the MMD with some predefined models \cite{Dann1, DDC, TCA, SPDA, TSC}, while some recent approaches were proposed to further study the MMD in-depth and elaborately improved the MMD using some prior data information \cite{BDA, MEDA, WMMD, DAN, JAN, HoMM}. However, the adverse impacts of MMD on some domain-specific data properties that are implicit in the original feature space are ignored carelessly, such as the feature discriminability and the local manifold structure of data, etc. Differently, some remarkable algorithms were raised to accumulate to devise various regularization losses independently of the MMD to offset the negative influences that arise from the MMD \cite{VDA, DICD, TIT, GEF, LPJT, TRSC, ARTL, SCA, JGSA}. Especially, some discriminative DA methods established the discriminative distances (i.e., intra-class and inter-class distances) like Local Discriminant Analysis (LDA) \cite{HMLDA} to further respect the discriminative information of each specific domain. However, it is hard to study the relationships between those statistics qualitatively and quantitatively due to deficiencies in the theory, so that more parameters have to be estimated blindly and the learned DA models are often unstable. Strikingly, this paper mainly aims to answer the following two essential questions: 1) how exactly does the MMD minimize the mean deviations between the two different domains? 2) why does the MMD usually produce an unexpected deterioration of the feature discriminability?

\par As we all known, the DA technique, as a branch of transfer learning, also aimed to simulate the transferable intelligence of human beings, so that the feature representations of those images with the same semantic (i.e., category) are as similar as possible. As shown in Fig. \ref{fig1}, we expect to leverage the common features shared between the source and target domains by minimizing their mean deviations of each pair of categories, even though they follow very different distributions, but how exactly does it work? Strikingly, this paper presents a novel insight into the MMD and theoretically reveals that its working principle could reach a high consensus with the transferable behavior of human beings. As shown in Fig. \ref{fig1}, given a pair of classes with the label Desktop\_computer respectively from the source and target domains, 1) the two relatively smaller red circles (i.e., the hollow and meshed ones) are transformed into the larger red ones which are magnified greatly (i.e., maximizing their specific intra-class distances); 2) the two tiny red circles are gradually drawn closer to each other along with their specific arrows (i.e., minimizing their joint variance). It is generally known that human beings attempt to abstract the common feature of a provided semantic, so that it could embrace all its possible appearances, i.e., 1), but the detailed information is decayed heavily, i.e., 2). Therefore, the proposed novel insight is highly consistent with the transferability of human beings, and the detailed theoretical proof will be elaborated in the Section \ref{a Novel Insight of the Class-Wise Maximum Mean Discrepancy}. Remarkably, this paper theoretically proves that the discriminative distances involved in the MMD are distinctly different from the ones in the LDA \cite{HMLDA} since there exist different weights imposed on various classes. As shown in Fig. \ref{fig1}, those relatively smaller circles are enlarged and drawn closer with varying degrees.  

\par Given the facts illustrated above, the reasons for degradation of the feature discriminability in the MMD could be therefore revealed. As can be seen from Fig. \ref{fig1}, the common features of Desktop\_computer and Trash\_can (resp. Laptop\_computer and Monitor) are quite similar, and the feature discriminability performs worse than before since the circles of different categories will chaotically overlap to each other. This observation provides us qualitative and quantitative guidance in studying the relationship between the MMD and the discriminative distances (i.e., transferability vs discriminability), and devising more robust and effective DA models. Qualitatively, this paper proposes a discriminative MMD with two different strategies to prompt the leveraged features more discriminative: 1) we directly impose a trade-off parameter on the implicit intra-class distance of the MMD to regulate its change; 2) we impose the similar weights revealed in the MMD on the inter-class distance and maximize it. Although two parameters are still involved, we have no longer need to estimate it blindly in the unknown regions and it is easy to know how the feature properties will change with varying parameter values since we experimentally observe that the implicit weights imposed on the discriminative distances revealed here are very close to the optimal ones empirically set in existing discriminative DA models (Section \ref{Ablation Study}), which theoretically gives the approximate optimal parameter ranges, thus different feature properties could be regulated more exactly. Specifically, the trade-off parameter in the first strategy could be set between $-1$ and $1$, where the intra-class distance is regulated corresponding with different physical meanings (i.e., suppressing/removing its expansion, enhancing its compactness instead). Moreover, the balanced factor in the second manner could be set between $0$ and $1$, to quantitatively leverage the relative importance between the feature transferability and its discriminability. Finally, we consider the intra-class and inter-class distances alone in the proposed two strategies, since this paper proves that their relationship is as one falls, another rises, thus the redundant parameter could be omitted. By and large, the main contributions of our work are three-folds: 

\begin{itemize}
	\item This paper theoretically proves that the working principle of the MMD, and illustrate the reasons for degradation of feature discriminability, which provides qualitative and quantitative guidance in studying their relationship and devising more robust and effective DA models.
	\item This paper proposes a discriminative MMD with two different strategies, which provides the approximate optimal parameter ranges and the corresponding physical meanings with different parameter values, thus the different feature properties could be regulated more exactly.
	\item This paper considers the intra-class and inter-class distances alone in the proposed two strategies since we prove that their relationship is as one falls, another rises, thus the redundant parameter could be further omitted.
\end{itemize}
 


\section{Rethink MMD}

\subsection{Preliminary}

\par In this paper, a matrix is denoted as the bold-italic uppercase letter (e.g., $\textbf{\textit{X}}$) but a column vector is represented as the bold-italic lowercase letter (e.g., $\textbf{\textit{x}}$), and $\textbf{\textit{x}}_i$ is the $i$-th column of $\textbf{\textit{X}}$. In addition, $(\textbf{\textit{X}})_{ij}$ is the value from the $i$-th row and the $j$-th column of $\textbf{\textit{X}}$, and $(\textbf{\textit{x}}_i)_i$ is the $i$-th value of $\textbf{\textit{x}}_i$. $||\textbf{\textit{X}}||_F^2=\sum_{i}\sum_{j}(\textbf{\textit{X}})_{ij}^2$ is the Frobenius norm of $\textbf{\textit{X}}$, and $||\textbf{\textit{x}}_i||_2^2=\sum_{j}(\textbf{\textit{x}}_i)_j^2$ is the $l_2$-norm of $\textbf{\textit{x}}_i$. The superscript $\top$ and $tr(\bullet)$ are the transpose and trace operators. The indexes of source and target domains are denoted using the subscripts $s$ and $t$. The index of the $c$-th category is defined by the superscript $c$. 
  
\par In DA scenario, there have a labled source domain (i.e., data matrix $\textbf{\textit{X}}_s\in \mathbf{R}^{m\times n_s}$, label vector $\textbf{\textit{y}}_s\in \mathbf{R}^{n_s}$) but an unlabeled target domain (i.e., $\textbf{\textit{X}}_t\in \mathbf{R}^{m\times n_t}$), where $m$ is the feature dimension and $n_s/n_t$ is the number of source/target data instances, and the whole data matrix is $\textbf{\textit{X}}_{st}=[\textbf{\textit{X}}_s, \textbf{\textit{X}}_t]\in \mathbf{R}^{m\times n_{st}}$ ($n_{st}=n_s+n_t$). Notably, if not stated otherwise, $\textbf{\textit{X}}\in \mathbf{R}^{m\times n}$ represents the data matrix of any given domain, and its label vector is $\textbf{\textit{y}}\in \mathbf{R}^n$. Our goal is to jointly project the source and target domains data into a common feature subspace so that the distributional differences between their new feature representations (i.e., $\textbf{\textit{Z}}_s=\textbf{\textit{A}}^{\top}\textbf{\textit{X}}_s\in \mathbf{R}^{k\times n_s}, \textbf{\textit{Z}}_t=\textbf{\textit{A}}^{\top}\textbf{\textit{X}}_t\in \mathbf{R}^{k\times n_t}, \textbf{\textit{A}}\in \mathbf{R}^{m\times k}$) are minimized substantially.

\subsection{Revisit MMD}

\par To be specific, the marginal distribution difference between the two domains could be measured using the Maximum Mean Discrepancy (MMD), which computes the deviation of their specific means, and could be formulated as follows

\begin{equation}
||\frac{1}{n_s}\sum_{\textbf{\textit{x}}_i\in \textbf{\textit{X}}_s}\textbf{\textit{A}}^{\top}\textbf{\textit{x}}_i-\frac{1}{n_t}\sum_{\textbf{\textit{x}}_j\in \textbf{\textit{X}}_t}\textbf{\textit{A}}^{\top}\textbf{\textit{x}}_j||_2^2=tr(\textbf{\textit{A}}^{\top}\textbf{\textit{X}}_{st}\textbf{\textit{M}}_0\textbf{\textit{X}}_{st}^{\top}\textbf{\textit{A}}),
\label{eq1}
\end{equation}

\noindent where $\textbf{\textit{M}}_0\in \mathbf{R}^{n_{st}\times n_{st}}$ is computed as follows

\begin{equation} 
(\textbf{\textit{M}}_0)_{ij}=\left\{ 
\begin{array}{lr}

\frac{1}{n_sn_s},  (\textbf{\textit{x}}_i,\textbf{\textit{x}}_j\in {\textbf{\textit{X}}_s}) &  \\

\frac{1}{n_tn_t},  (\textbf{\textit{x}}_i,\textbf{\textit{x}}_j\in {\textbf{\textit{X}}_t}) &  \\

-\frac{1}{n_sn_t}, (otherwise). &  

\end{array}
\right .
\label{eq2}
\end{equation}

\noindent Together with label constraint, the class-wise MMD is modeled to approximately measure the conditional distribution difference across the two domains, which further computes the mean deviations of each two coupled classes with the same labels but from different domains, and it could be defined as follows

\begin{equation}
\begin{array}{lr}
\sum_{c=1}^{C}||\frac{1}{n_s^c}\sum_{\textbf{\textit{x}}_i\in \textbf{\textit{X}}_s^c}\textbf{\textit{A}}^{\top}\textbf{\textit{x}}_i-\frac{1}{n_t^c}\sum_{\textbf{\textit{x}}_j\in \textbf{\textit{X}}_t^c}\textbf{\textit{A}}^{\top}\textbf{\textit{x}}_j||_2^2 \\
=\sum_{c=1}^{C}tr(\textbf{\textit{A}}^{\top}\textbf{\textit{X}}_{st}\textbf{\textit{M}}_c\textbf{\textit{X}}_{st}^{\top}\textbf{\textit{A}}),
\end{array}
\label{eq3}
\end{equation}

\noindent where $\textbf{\textit{X}}_s^c$ (resp. $\textbf{\textit{X}}_t^c$) is the data samples pertaining to the $c$-th category of source domain (resp. target domain), and its number is $n_s^c$ (resp. $n_t^c$). Likewise, the $\textbf{\textit{M}}_c\in \mathbf{R}^{n_{st}\times n_{st}}$ is computed as follows

\begin{equation}
(\textbf{\textit{M}}_c)_{ij}=\left \{ 
\begin{array}{lr}

\frac{1}{n_s^cn_s^c},  (\textbf{\textit{x}}_i,\textbf{\textit{x}}_j\in {\textbf{\textit{X}}_s^c}) &  \\

\frac{1}{n_t^cn_t^c},  (\textbf{\textit{x}}_i,\textbf{\textit{x}}_j\in {\textbf{\textit{X}}_t^c}) &  \\

-\frac{1}{n_s^cn_t^c}, \left \{ \begin{array}{lr}
\textbf{\textit{x}}_i\in {\textbf{\textit{X}}_s^c}, \textbf{\textit{x}}_j \in{\textbf{\textit{X}}_t^c}  \\ \textbf{\textit{x}}_j\in {\textbf{\textit{X}}_s^c}, \textbf{\textit{x}}_i\in {\textbf{\textit{X}}_t^c} \end{array} \right.    & \\

0, (otherwise). &  

\end{array}
\right.
\label{eq4}
\end{equation}

\par Notably, the MMD is a special case of class-wise MMD where the whole source/target domain data is regarded as one category. Once those two metrics of distributional distance are established, we can jointly reduce the marginal and conditional distribution differences between the source and target domains by minimizing the MMD and class-wise MMD losses in a given feature learning framework (e.g., the Principal Component Analysis, PCA), and it could be formulated as follows

\begin{equation}
\begin{array}{lr}
\min\limits_{\textbf{\textit{A}}}\sum_{c=0}^{C}tr(\textbf{\textit{A}}^{\top}\textbf{\textit{X}}_{st}\textbf{\textit{M}}_c\textbf{\textit{X}}_{st}^{\top}\textbf{\textit{A}})+\alpha||\textbf{\textit{A}}||_F^2 \\ 
\qquad s.t. \quad \textbf{\textit{A}}^{\top}\textbf{\textit{X}}_{st}\textbf{\textit{H}}_{st}\textbf{\textit{X}}_{st}^{\top}\textbf{\textit{A}}=\textbf{\textit{I}}_{k\times k},
\end{array}
\label{eq5}
\end{equation}

\noindent where the constraint $\textbf{\textit{A}}^{\top}\textbf{\textit{X}}_{st}\textbf{\textit{H}}_{st}\textbf{\textit{X}}_{st}^{\top}\textbf{\textit{A}}=\textbf{\textit{I}}_{k\times k}$ means that the whole data variance is tied to a fixed value, so that the data information on the subspace could be statistically preserved to some extent. $||\textbf{\textit{A}}||_F^2$ controls the scale of projection $\textbf{\textit{A}}$, and $\alpha$ is the trade-off parameter. Notably, $\textbf{\textit{I}}_{\bullet\times \bullet}$ denotes an identity matrix with the size of $\bullet \times \bullet$, $\textbf{\textit{1}}_{\bullet \times \bullet}$ is a matrix whose elements are all $1$ with the size of $\bullet \times \bullet$, and $\textbf{\textit{H}}_{st}=\textbf{\textit{I}}_{n_{st}\times n_{st}}-\frac{1}{n_{st}}\textbf{\textit{1}}_{n_{st}\times n_{st}}$ is a centering matrix.

\subsection{Rethink MMD}
\label{a Novel Insight of the Class-Wise Maximum Mean Discrepancy}
\par Although the MMD has been widely utilized in the cross-domain problem, its detailed working principle is still under insufficient exploration so far. Notably, this paper only focuses on the class-wise MMD because it plays a decisive role in the DA performances, and the MMD is a special case of class-wise MMD.

\par This paper mainly answers the following two essential questions: 1) how exactly does the MMD minimize the mean deviations between the two different domains? 2) why does the MMD usually produce an unexpected deterioration of the feature discriminability? To this end, a novel insight is firstly provided theoretically, and we now present Lemma 1 $\sim$ Lemma 3 as follows

\noindent \textbf{Lemma 1.} We have the following identity about the inter-class distance according to \cite{HMLDA} 
\begin{equation}
tr(\textbf{\textit{A}}^{\top}\textbf{\textit{S}}_b\textbf{\textit{A}})=\frac{1}{n}\sum_{i=1}^{C-1}\sum_{j=i+1}^Cn^in^jtr(\textbf{\textit{A}}^{\top}\textbf{\textit{D}}^{ij}\textbf{\textit{A}}),
\label{eq6}
\end{equation}

\noindent where $\textbf{\textit{D}}^{ij}=(\textbf{\textit{m}}^i-\textbf{\textit{m}}^j)(\textbf{\textit{m}}^i-\textbf{\textit{m}}^j)^{\top}$, and $\textbf{\textit{S}}_b=\sum_{i=1}^Cn^i(\textbf{\textit{m}}^i-\textbf{\textit{m}})(\textbf{\textit{m}}^i-\textbf{\textit{m}})^{\top}$ is the inter-class scatter matrix. Notably, $\textbf{\textit{m}}^i$/$\textbf{\textit{m}}^j$ denotes the mean of data instances from the $i$/$j$-th category, the number of data instances is $n^i/n^j$, and $\textbf{\textit{m}}$ is the mean of the whole data samples.
\\
\noindent \textbf{Lemma 2.} The inter-class distance equals to the data variance minus the intra-class distance
\begin{equation}
tr(\textbf{\textit{A}}^{\top}\textbf{\textit{S}}_b\textbf{\textit{A}})=tr(\textbf{\textit{A}}^{\top}\textbf{\textit{S}}_v\textbf{\textit{A}})-tr(\textbf{\textit{A}}^{\top}\textbf{\textit{S}}_w\textbf{\textit{A}}),
\label{eq7}
\end{equation}

\noindent where $\textbf{\textit{S}}_v=\sum_{i=1}^n(\textbf{\textit{x}}_i-\textbf{\textit{m}})(\textbf{\textit{x}}_i-\textbf{\textit{m}})^{\top}$ is the variance matrix, and $\textbf{\textit{S}}_w=\sum_{i=1}^C\sum_{\textbf{\textit{x}}_j\in \textbf{\textit{X}}^i}(\textbf{\textit{x}}_j-\textbf{\textit{m}}^i)(\textbf{\textit{x}}_j-\textbf{\textit{m}}^i)^{\top}$ is the intra-class scatter matrix.

\noindent \textbf{Proof.} $\textbf{\textit{S}}_w+\textbf{\textit{S}}_b=\sum_{i=1}^C\sum_{\textbf{\textit{x}}_j\in\textbf{\textit{X}}^i}(\textbf{\textit{x}}_j-\textbf{\textit{m}}^i)(\textbf{\textit{x}}_j-\textbf{\textit{m}}^i)^{\top}+(\textbf{\textit{m}}^i-\textbf{\textit{m}})(\textbf{\textit{m}}^i-\textbf{\textit{m}})^{\top}=\sum_{i=1}^C\sum_{\textbf{\textit{x}}_j\in\textbf{\textit{X}}^i}\textbf{\textit{x}}_j\textbf{\textit{x}}_j^{\top}-2\textbf{\textit{x}}_j\textbf{\textit{m}}_i^{\top}+\textbf{\textit{m}}_i\textbf{\textit{m}}_i^{\top}+\textbf{\textit{m}}_i\textbf{\textit{m}}_i^{\top}-2\textbf{\textit{m}}_i\textbf{\textit{m}}^{\top}+\textbf{\textit{m}}\textbf{\textit{m}}^{\top}=\sum_{i=1}^C\sum_{\textbf{\textit{x}}_j\in\textbf{\textit{X}}^i}\textbf{\textit{x}}_j\textbf{\textit{x}}_j^{\top}-2(\textbf{\textit{x}}_j\textbf{\textit{m}}_i^{\top}-\textbf{\textit{m}}_i\textbf{\textit{m}}_i^{\top}+\textbf{\textit{m}}_i\textbf{\textit{m}}^{\top})+\textbf{\textit{m}}\textbf{\textit{m}}^{\top}=\sum_{i=1}^n(\textbf{\textit{x}}_i\textbf{\textit{x}}_i^{\top}+\textbf{\textit{m}}\textbf{\textit{m}}^{\top})-2\sum_{i=1}^C\sum_{\textbf{\textit{x}}_j\in\textbf{\textit{X}}^i}(\textbf{\textit{x}}_j\textbf{\textit{m}}_i^{\top}-\textbf{\textit{m}}_i\textbf{\textit{m}}_i^{\top}+\textbf{\textit{m}}_i\textbf{\textit{m}}^{\top})$. \\
because $\sum_{i=1}^C\sum_{\textbf{\textit{x}}_j\in\textbf{\textit{X}}^i}(\textbf{\textit{x}}_j\textbf{\textit{m}}_i^{\top}-\textbf{\textit{m}}_i\textbf{\textit{m}}_i^{\top})=\sum_{i=1}^C((\textbf{\textit{x}}_1+...+\textbf{\textit{x}}_{n^i})\textbf{\textit{m}}_i^{\top}-n^i\textbf{\textit{m}}_i\textbf{\textit{m}}_i^{\top})$ and $n^i\textbf{\textit{m}}_i=\textbf{\textit{x}}_1+...+\textbf{\textit{x}}_{n^i}$. Then $\sum_{i=1}^C\sum_{\textbf{\textit{x}}_j\in\textbf{\textit{X}}^i}(\textbf{\textit{x}}_j\textbf{\textit{m}}_i^{\top}-\textbf{\textit{m}}_i\textbf{\textit{m}}_i^{\top})=0$, and $\textbf{\textit{S}}_w+\textbf{\textit{S}}_b=\sum_{i=1}^n(\textbf{\textit{x}}_i\textbf{\textit{x}}_i^{\top}+\textbf{\textit{m}}\textbf{\textit{m}}^{\top}-2\textbf{\textit{m}}_i\textbf{\textit{m}}^{\top})=\sum_{i=1}^n(\textbf{\textit{x}}_i-\textbf{\textit{m}})(\textbf{\textit{x}}_i-\textbf{\textit{m}})^{\top}=\textbf{\textit{S}}_v$. This completes the proof.
\\
\noindent \textbf{Lemma 3.} We have the following identity about the MMD
\begin{equation}
\begin{array}{lr}
\sum_{c=1}^Ctr(\textbf{\textit{A}}^{\top}\textbf{\textit{X}}\textbf{\textit{M}}_c\textbf{\textit{X}}^{\top}\textbf{\textit{A}})=\sum_{c=1}^C\frac{n_s^c+n_t^c}{n_s^cn_t^c}tr(\textbf{\textit{A}}^{\top}(\textbf{\textit{S}}_{st})_b^c\textbf{\textit{A}}) \\
=\sum_{c=1}^C\frac{n_s^c+n_t^c}{n_s^cn_t^c}tr(\textbf{\textit{A}}^{\top}(\textbf{\textit{S}}_{st})_v^c\textbf{\textit{A}}) \\
\qquad \qquad \qquad \qquad \qquad -\sum_{c=1}^C\frac{n_s^c+n_t^c}{n_s^cn_t^c}tr(\textbf{\textit{A}}^{\top}(\textbf{\textit{S}}_{st})_w^c\textbf{\textit{A}}),
\end{array}
\label{eq8}
\end{equation} 

\noindent where $(\textbf{\textit{S}}_{st})_b^c=\sum_{i\in \{s,t\}}n_i^c(\textbf{\textit{m}}_i^c-\textbf{\textit{m}}_{st}^c)(\textbf{\textit{m}}_i^c-\textbf{\textit{m}}_{st}^c)^{\top}$, $(\textbf{\textit{S}}_{st})_v^c=\sum_{i=1}^{n_{st}^c}(\textbf{\textit{x}}_i-\textbf{\textit{m}}_{st}^c)(\textbf{\textit{x}}_i-\textbf{\textit{m}}_{st}^c)^{\top}$, and $(\textbf{\textit{S}}_{st})_w^c=\sum_{i\in \{s,t\}}\sum_{j=1}^{n_i^c}(\textbf{\textit{x}}_j-\textbf{\textit{m}}_i^c)(\textbf{\textit{x}}_j-\textbf{\textit{m}}_i^c)^{\top}$. Notably, $n_i^c$ is the number of data instances of the $c$-th category from the $i$-th domain, and $n_{st}^c=n_s^c+n_t^c$. Besides, $\textbf{\textit{m}}_i^c$ is the data mean of the $c$-th category from the $i$-th domain, and $\textbf{\textit{m}}_{st}^c$ is the data mean of the $c$-th category from both the source and target domains. 
\\
\noindent \textbf{Proof.} Given the data instances of $c$-th categories respectively from the source and target domains, because $tr(\textbf{\textit{A}}^{\top}\textbf{\textit{X}}\textbf{\textit{M}}_c\textbf{\textit{X}}^{\top}\textbf{\textit{A}})=tr(\textbf{\textit{A}}^{\top} \\
(\textbf{\textit{m}}_s^c-\textbf{\textit{m}}_t^c)(\textbf{\textit{m}}_s^c-\textbf{\textit{m}}_t^c)^{\top}\textbf{\textit{A}})=tr(\textbf{\textit{A}}^{\top}
\textbf{\textit{D}}^{s^ct^c}\textbf{\textit{A}})$, and Lemma 1, Lemma 2, we have
\begin{equation}
\begin{array}{lr}
tr(\textbf{\textit{A}}^{\top}\textbf{\textit{X}}\textbf{\textit{M}}_c\textbf{\textit{X}}^{\top}\textbf{\textit{A}})=\frac{n_s^c+n_t^c}{n_s^cn_t^c}tr(\textbf{\textit{A}}^{\top}(\textbf{\textit{S}}_{st})_b^c\textbf{\textit{A}}) \\
=\frac{n_s^c+n_t^c}{n_s^cn_t^c}tr(\textbf{\textit{A}}^{\top}(\textbf{\textit{S}}_{st})_v^c\textbf{\textit{A}})-\frac{n_s^c+n_t^c}{n_s^cn_t^c}tr(\textbf{\textit{A}}^{\top}(\textbf{\textit{S}}_{st})_w^c\textbf{\textit{A}}).
\end{array}
\label{eq9}
\end{equation} 
\noindent Notably, the Eq. \ref{eq8} is the sum of Eq. \ref{eq9}. This completes the proof.

\par From Eq. \ref{eq9}, it could be concluded that the MMD aims to respectively maximize the source and target intra-class distances but jointly minimize their variance with different weights, which are separately established by each pair of classes with the same labels but from the source and target domains. This conclusion illustrates the detailed working principle of the MMD that how it precisely minimizes the mean deviations. Moreover, since the intra-class distance is enlarged greatly and the whole data variance is tied to a fixed value (i.e., Eq. \ref{eq5}), the inter-class distance will be drawn closer (i.e., Lemma 2). Thus, different classes will chaotically overlap to each other with various degrees, and the feature discriminability is degraded largely, which provides us qualitative and quantitative guidance in studying the relationship between the MMD and the discriminative distances, and devising more robust and effective DA models. Based on this, in the next subsection, we will propose a discriminative MMD with two different strategies to prompt the extracted features more discriminative. 

\section{The Proposed Approach}

\par According to Lemma 3, we could rewrite Eq. \ref{eq5} using the equivalent formulation of the original MMD as follows

\begin{equation}
\begin{array}{lr}
\min\limits_{\textbf{\textit{A}}}tr(\textbf{\textit{A}}^{\top}\textbf{\textit{X}}_{st}\textbf{\textit{M}}_0\textbf{\textit{X}}_{st}^{\top}\textbf{\textit{A}})+\sum_{c=1}^{C}w_{st}^ctr(\textbf{\textit{A}}^{\top}\textbf{\textit{X}}_{st}((\textbf{\textit{L}}_{st})_v^c
\\
-(\textbf{\textit{L}}_{st})_w^c)
\textbf{\textit{X}}_{st}^{\top}\textbf{\textit{A}})+\alpha||\textbf{\textit{A}}||_F^2 \quad s.t. \quad \textbf{\textit{A}}^{\top}\textbf{\textit{X}}_{st}\textbf{\textit{H}}_{st}\textbf{\textit{X}}_{st}^{\top}\textbf{\textit{A}}=\textbf{\textit{I}}_{k\times k},
\end{array}
\label{eq10}
\end{equation}

\noindent where $w_{st}^c=\frac{n_{st}^c}{n_s^cn_t^c}$, and $(\textbf{\textit{L}}_{st})_v^c$ and $(\textbf{\textit{L}}_{st})_w^c$ are the Laplacian matrix of $(\textbf{\textit{S}}_{st})_v^c$ and $(\textbf{\textit{S}}_{st})_w^c$, respectively. We let $(\textbf{\textit{L}}_{st})_v^{c*}=\textbf{\textit{I}}_{n_{st}^c}-\frac{1}{n_{st}^c}\textbf{\textit{1}}_{n_{st}^c\times n_{st}^c}$, and define $\textbf{\textit{V}}_{st}^{c*}\in \mathbf{R}^{n_{st}^c\times n_{st}^c}$ as follows

\begin{equation}
(\textbf{\textit{V}}_{st}^{c*})_{ij}=\left\{ 
\begin{array}{lr}
\frac{1}{n_{s}^cn_{s}^c}, (\textbf{\textit{x}}_i,\textbf{\textit{x}}_j\in \textbf{\textit{X}}_s^c) &  \\
\frac{1}{n_{t}^cn_{t}^c}, (\textbf{\textit{x}}_i,\textbf{\textit{x}}_j\in \textbf{\textit{X}}_t^c) &  \\
0, (otherwise). &  

\end{array}
\right .
\label{eq11}
\end{equation} 

\noindent Moreover, we define $(G_{st}^{c*})_i=\sum_{j}(\textbf{\textit{V}}_{st}^{c*})_{ij}$ and $\textbf{\textit{G}}_{st}^{c*}=diag((G_{st}^{c*})_1,...,
\\
(G_{st}^{c*})_{n_{st}^c})$, thus $(\textbf{\textit{L}}_{st})_w^{c*}=\textbf{\textit{V}}_{st}^{c*}-\textbf{\textit{G}}_{st}^{c*}$. For the convenience of matrix operation in Eq. \ref{eq10}, we utilize $(\textbf{\textit{L}}_{st})_v^{c*}$ and $(\textbf{\textit{L}}_{st})_w^{c*}$ to define $(\textbf{\textit{L}}_{st})_v^c\in \mathbf{R}^{n_{st}\times n_{st}}$ and $(\textbf{\textit{L}}_{st})_w^c\in \mathbf{R}^{n_{st}\times n_{st}}$ as follows

\begin{equation}
\left\{ 
\begin{array}{lr}
((\textbf{\textit{L}}_{st})_v^{c})_{(\textbf{\textit{y}}_s=c,\textbf{\textit{y}}_s=c)}
\\ \qquad \qquad \qquad \qquad
=(\textbf{\textit{L}}_{st})_v^{c*}(1:n_s^c,1:n_s^c), &  \\
((\textbf{\textit{L}}_{st})_v^{c})_{(\textbf{\textit{y}}_t=c,\textbf{\textit{y}}_t=c)}
\\ \qquad \quad
=(\textbf{\textit{L}}_{st})_v^{c*}(n_s^c+1:n_s^c+n_t^c,n_s^c+1:n_s^c+n_t^c), &  \\
((\textbf{\textit{L}}_{st})_v^{c})_{(\textbf{\textit{y}}_s=c,\textbf{\textit{y}}_t=c)}
\\ \qquad \qquad \qquad \qquad
=(\textbf{\textit{L}}_{st})_v^{c*}(1:n_s^c,n_s^c+1:n_s^c+n_t^c), &  \\
((\textbf{\textit{L}}_{st})_v^{c})_{(\textbf{\textit{y}}_t=c,\textbf{\textit{y}}_s=c)}
\\ \qquad \qquad \qquad \qquad
=(\textbf{\textit{L}}_{st})_v^{c*}(n_s^c+1:n_s^c+n_t^c,1:n_s^c), &  \\
0, (otherwise). &  

\end{array}
\right .
\label{eq12}
\end{equation}

\begin{equation}
\left\{ 
\begin{array}{lr}
((\textbf{\textit{L}}_{st})_w^{c})_{(\textbf{\textit{y}}_s=c,\textbf{\textit{y}}_s=c)}
\\ \qquad \qquad \qquad \qquad
=(\textbf{\textit{L}}_{st})_w^{c*}(1:n_s^c,1:n_s^c), &  \\
((\textbf{\textit{L}}_{st})_w^{c})_{(\textbf{\textit{y}}_t=c,\textbf{\textit{y}}_t=c)}
\\ \qquad \quad
=(\textbf{\textit{L}}_{st})_w^{c*}(n_s^c+1:n_s^c+n_t^c,n_s^c+1:n_s^c+n_t^c), &  \\
((\textbf{\textit{L}}_{st})_w^{c})_{(\textbf{\textit{y}}_s=c,\textbf{\textit{y}}_t=c)}
\\ \qquad \qquad \qquad \qquad
=(\textbf{\textit{L}}_{st})_w^{c*}(1:n_s^c,n_s^c+1:n_s^c+n_t^c), &  \\
((\textbf{\textit{L}}_{st})_w^{c})_{(\textbf{\textit{y}}_t=c,\textbf{\textit{y}}_s=c)}
\\ \qquad \qquad \qquad \qquad
=(\textbf{\textit{L}}_{st})_w^{c*}(n_s^c+1:n_s^c+n_t^c,1:n_s^c), &  \\
0, (otherwise). &  

\end{array}
\right .
\label{eq13}
\end{equation}


\par Now, we devise a discriminative MMD with two different strategies, and they could be formulated as Eq. \ref{eq14} and Eq. \ref{eq15}. 

\subsection{The First Strategy}
\label{first}
\begin{equation}
\begin{array}{lr}
\min\limits_{\textbf{\textit{A}}}\sum_{c=1}^{C}\textbf{\textit{w}}_{st}^ctr(\textbf{\textit{A}}^{\top}\textbf{\textit{X}}_{st}((\textbf{\textit{L}}_{st})_v^c+\beta(\textbf{\textit{L}}_{st})_w^c)\textbf{\textit{X}}_{st}^{\top}\textbf{\textit{A}})
\\
+tr(\textbf{\textit{A}}^{\top}\textbf{\textit{X}}_{st}\textbf{\textit{M}}_0\textbf{\textit{X}}_{st}^{\top}\textbf{\textit{A}})+\alpha||\textbf{\textit{A}}||_F^2 
\\
\qquad \qquad \qquad \qquad s.t. \quad \textbf{\textit{A}}^{\top}\textbf{\textit{X}}_{st}\textbf{\textit{H}}_{st}\textbf{\textit{X}}_{st}^{\top}\textbf{\textit{A}}=\textbf{\textit{I}}_{k\times k}.
\end{array}
\label{eq14}
\end{equation}

\noindent Here, a trade-off parameter $\beta$ between $-1$ and $1$ is directly imposed on the implicit intra-class distance of MMD. Notably, the revealed weights could provide us the theoretical guidance for setting parameter imposed on the intra-class distance, and intensifying the feature discriminability more correctly, since the optimal parameter regions are revealed by the implicit weights, and we exactly knew how it will change the leveraged feature properties with varying $\beta$.   

\par Specifically, there exist three cases respecting Eq. \ref{eq14}: 1) the expansion of intra-class distance is gradually mitigated when $\beta\in [-1,0)$; 2) the adverse influences on the intra-class distance is offset exactly when $\beta=0$; 3) the intra-class compactness is positively stimulated instead of weakening it when $\beta\in (0,1]$.

\par Similar to previous work \cite{JDA, BDA, VDA}, Eq. \ref{eq14} is equivalent to a generalized eigen-decomposition problem as follows

\begin{equation}
\begin{array}{lr}
(\sum_{c=1}^{C}\textbf{\textit{X}}_{st}(\textbf{\textit{w}}_{st}^c(\textbf{\textit{L}}_{st})_v^c-\textbf{\textit{w}}_{st}^c\beta(\textbf{\textit{L}}_{st})_w^c+\textbf{\textit{M}}_0)\textbf{\textit{X}}_{st}^{\top}+\alpha\textbf{\textit{I}}_{m\times m})\textbf{\textit{A}}
\\
=\textbf{\textit{X}}_{st}\textbf{\textit{H}}_{st}\textbf{\textit{X}}_{st}^{\top}\textbf{\textit{A}}\boldsymbol{\Theta},
\end{array}
\label{eq15}
\end{equation}

\noindent where $\boldsymbol{\Theta}\in \mathbf{R}^{k\times k}$ is a diagonal matrix with Lagrange Multipliers. The Eq. \ref{eq15} can be effectively and efficiently solved by calculating the eigenvectors corresponding to the $k$-smallest eigenvalues.

\subsection{The Second Strategy}
\label{second} 
\begin{equation}
\begin{array}{lr}
\min\limits_{\textbf{\textit{A}}}\lambda\sum_{c=1}^{C}\textbf{\textit{w}}_{st}^ctr(\textbf{\textit{A}}^{\top}\textbf{\textit{X}}_{st}((\textbf{\textit{L}}_{st})_v^c-(\textbf{\textit{L}}_{st})_w^c)\textbf{\textit{X}}_{st}^{\top}\textbf{\textit{A}})
\\
-(1-\lambda)\sum_{i=1}\sum_{j=i+1}tr(\textbf{\textit{A}}^{\top}\textbf{\textit{X}}_{st}[(\textbf{\textit{L}}_s)_b^{ij}, \textbf{\textit{0}};\textbf{\textit{0}}, (\textbf{\textit{L}}_t)_b^{ij}]\textbf{\textit{X}}_{st}^{\top}\textbf{\textit{A}})
\\
+tr(\textbf{\textit{A}}^{\top}\textbf{\textit{X}}_{st}\textbf{\textit{M}}_0\textbf{\textit{X}}_{st}^{\top}\textbf{\textit{A}})+\alpha||\textbf{\textit{A}}||_F^2 
\\
\qquad \qquad \qquad \qquad s.t. \quad \textbf{\textit{A}}^{\top}\textbf{\textit{X}}_{st}\textbf{\textit{H}}_{st}\textbf{\textit{X}}_{st}^{\top}\textbf{\textit{A}}=\textbf{\textit{I}}_{k\times k},
\end{array}
\label{eq16}
\end{equation}

\noindent where $(\textbf{\textit{L}}_s)_b^{ij}=\textbf{\textit{w}}_s^{ij}((\textbf{\textit{L}}_s)_v^{ij}-(\textbf{\textit{L}}_s)_w^{ij})$, $(\textbf{\textit{L}}_t)_b^{ij}=\textbf{\textit{w}}_t^{ij}((\textbf{\textit{L}}_t)_v^{ij}-(\textbf{\textit{L}}_t)_w^{ij})$ are the Laplacian matrix, $\textbf{\textit{w}}_s^{ij}=\frac{n_s^{ij}}{n_s^i+n_s^j}$, $\textbf{\textit{w}}_t^{ij}=\frac{n_t^{ij}}{n_t^i+n_t^j}$, and $n_s^{ij}=n_s^i+n_s^j, n_t^{ij}=n_t^i+n_t^j$. Similarly, $(\textbf{\textit{L}}_{s/t})_v^{ij*}=\textbf{\textit{I}}_{n_{s/t}^{ij}}-\frac{1}{n_{s/t}^{ij}}\textbf{\textit{1}}_{n_{s/t}^{ij} \times n_{s/t}^{ij}}$, and $(\textbf{\textit{L}}_{s/t})_w^{ij*}=\textbf{\textit{V}}_{s/t}^{ij*}-\textbf{\textit{G}}_{s/t}^{ij*}$, where $(G_{s/t}^{ij*})_{h}=\sum_{h}(\textbf{\textit{V}}_{s/t}^{ij*})_{lh}$, $\textbf{\textit{G}}_{s/t}^{ij*}=diag((G_{s/t}^{ij*})_1,...,
(G_{s/t}^{ij*})_{n_{s/t}^{ij}})$, and $\textbf{\textit{V}}_{s/t}^{ij*}\in \mathbf{R}^{n_{s/t}^{ij}\times n_{s/t}^{ij}}$ could be computed as follows

\begin{equation}
(\textbf{\textit{V}}_{s/t}^{ij*})_{lh}=\left\{ 
\begin{array}{lr}
\frac{1}{n_{s/t}^{i}n_{s/t}^i}, (\textbf{\textit{x}}_l,\textbf{\textit{x}}_h\in \textbf{\textit{X}}_{s/t}^i) &  \\
\frac{1}{n_{s/t}^{j}n_{s/t}^j}, (\textbf{\textit{x}}_l,\textbf{\textit{x}}_h\in \textbf{\textit{X}}_{s/t}^j) &  \\
0, (otherwise). &  

\end{array}
\right .
\label{eq17}
\end{equation} 

\noindent Similar to $(\textbf{\textit{L}}_{st})_v^c$ and $(\textbf{\textit{L}}_{st})_w^c$, $(\textbf{\textit{L}}_{s/t})_v^{ij}$ and $(\textbf{\textit{L}}_{s/t})_w^{ij}$ could be constructed accordingly. Different from the first strategy, we aim to further reformulate the inter-class distances of the source and target domains using the similar weights revealed in the MMD so that the optimal parameter regions imposed on the inter-class distances are also known beforehand. Therefore, a balanced factor of $\lambda$ could be employed to adaptively leverage the importance of feature transferability and its discriminability. Likewise, Eq. \ref{eq15} could be solved in the same manner as Eq. \ref{eq14}.  
\par Moreover, from Lemma 3, we consider those two discriminative distances alone in the proposed two strategies since their relationship is as one falls, another rises when the whole data variance is fixed, thus the redundant parameter could be further omitted.    
   
\subsection{Classification Scheme}

\par We utilize the whole common features $\textbf{\textit{Z}}=[\textbf{\textit{Z}}_s,\textbf{\textit{Z}}_t]$ to construct a neighborhood similarity graph $\textbf{\textit{W}}\in \mathbf{R}^{n_{st}\times n_{st}}$ with the $p$-neareast neighbors similar to \cite{CPCAN,GAKT}. Specifically, the weight $(\textbf{\textit{W}})_{ij}$ mesures the similarity degree between $\textbf{\textit{x}}_i$ and $\textbf{\textit{x}}_j$, and the closer, the bigger. We now employ a Graph-based Label Propogation method \cite{GAKT} (GLP) to propagate the source labels to the target domain data as follows

\begin{equation}
\begin{array}{lr}
\min\limits_{\textbf{\textit{F}}_t}\sum_{i=1}^{n_{st}}\sum_{j=1}^{n_{st}}(\textbf{\textit{W}})_{ij}||\textbf{\textit{f}}_i-\textbf{\textit{f}}_j||_2^2
\\
=\min\limits_{\textbf{\textit{F}}_t}tr(\textbf{\textit{F}}_t\textbf{\textit{L}}^{tt}\textbf{\textit{F}}_t^{\top})+2tr(\textbf{\textit{F}}_t\textbf{\textit{L}}^{ts}\textbf{\textit{F}}_s^{\top}),
\end{array}
\label{eq18}
\end{equation}

\noindent where $\textbf{\textit{F}}_s\in \mathbf{R}^{C \times n_s}, \textbf{\textit{F}}_t\in \mathbf{R}^{C \times n_t}$, $\textbf{\textit{F}}=[\textbf{\textit{F}}_s,\textbf{\textit{F}}_t]\in \mathbf{R}^{C \times n_{st}}$ are the one-hot labels, and $(\textbf{\textit{F}})_{ci}=1$, $(\textbf{\textit{F}})_{li}=0, l\neq c$ if $(\textbf{\textit{y}})_i=c$. Besides, we define the graph Laplacian matrix $\textbf{\textit{L}}=\textbf{\textit{B}}-\textbf{\textit{W}}$, where $\textbf{\textit{B}}$ denotes a diagonal matrix with the diagonal entries as the column sums of $\textbf{\textit{W}}$, and $\textbf{\textit{L}}=[\textbf{\textit{L}}_{ss},\textbf{\textit{L}}_{st};\textbf{\textit{L}}_{ts},\textbf{\textit{L}}_{tt}]$.

\par We first obtain the partial derivative of Eq. \ref{eq17} w.r.t., $\textbf{\textit{F}}_t$, and set it to $\textbf{\textit{0}}$. Then the solution can be derived as follows

\begin{equation}
\textbf{\textit{F}}_t=\textbf{\textit{F}}_s(\textbf{\textit{L}}^{ts})^{\top}(\textbf{\textit{L}}^{tt})^{-1},
\label{eq19}
\end{equation}

\noindent Once the one-hot label matrix $\textbf{\textit{F}}_t$ is obtained, the target label of any given data instance $\textbf{\textit{x}}_i$ is computed as $(\textbf{\textit{y}}_t)_i=argmax_j(\textbf{\textit{F}}_t)_{ji}$.

\par Since the statistics about the target domain are computed by their pseudo labels during the current iteration, we have to optimize $\textbf{\textit{A}}$ and $\textbf{\textit{F}}_t$ iteratively. Remarkably, our approach could achieve desirable performances within only a few iterations (i.e., $T$).

\begin{table*}[!h]
	\caption{Average Classification Accuracy(\%) of Office-10 vs Caltech-10 with the SURF features}
	\label{tab1}
	\begin{tabular}{|c||c|c|c|c|c|c|c|c|c|c|c|c||c|}
		\hline
		Methods$/$Tasks & C$\rightarrow$A & C$\rightarrow$W & C$\rightarrow$D & A$\rightarrow$C & A$\rightarrow$W & A$\rightarrow$D & W$\rightarrow$C & W$\rightarrow$A & W$\rightarrow$D & D$\rightarrow$C & D$\rightarrow$A & D$\rightarrow$W & average \\
		\hline
		\hline
		JDA \cite{JDA}   & 45.62 & 41.69 & 45.22 & 39.36 & 37.97 & 39.49 & 31.17 & 32.78 & 89.17 & 31.52 & 33.09 & 89.49 & 46.38 \\  
		\hline
		BDA \cite{BDA}   & 44.89 & 38.64 & 47.77 & 40.78 & 39.32 & 43.31 & 28.94 & 32.99 & 91.72 & 32.50 & 33.09 & 91.86 & 47.15 \\
		\hline
		MEDA \cite{MEDA} & 56.50 & 53.90 & 50.30 & 43.90 & 53.20 & 45.90 & 34.00 & 42.70 & 88.50 & 34.90 & 41.20 & 87.50 & \underline{52.70} \\
		\hline
		ARTL \cite{ARTL} & 44.10 & 31.50 & 39.50 & 36.10 & 33.60 & 36.90 & 29.70 & 38.30 & 87.90 & 30.50 & 34.90 & 88.50 & 44.30 \\
		\hline
		VDA \cite{VDA}   & 46.14 & 46.10 & 51.59 & 42.21 & 51.19 & \textbf{48.41} & 27.60 & 26.10 & 89.18 & 31.26 & 37.68 & 90.85 & 49.03 \\
		\hline
		SCA \cite{SCA}   & 43.74 & 33.56 & 39.49 & 38.29 & 33.90 & 34.21 & 30.63 & 30.48 & 92.36 & 32.32 & 33.72 & 88.81 & 44.29 \\
		\hline
		JGSA \cite{JGSA} & 51.46 & 45.42 & 45.86 & 41.50 & 45.76 & 47.13 & 33.21 & 39.87 & 90.45 & 29.92 & 38.00 & 91.86 & 50.04 \\
		\hline
		DICD \cite{DICD} & 47.29 & 46.44 & 49.68 & 42.39 & 45.08 & 38.85 & 33.57 & 34.13 & 89.81 & 34.64 & 34.45 & 91.19 & 48.96 \\
		\hline
		TIT \cite{TIT}   & 59.70 & 51.50 & 48.40 & \textbf{47.50} & 45.40 & 47.10 & 34.90 & 40.20 & 87.90 & 36.70 & 42.10 & 84.80 & 52.20 \\
		\hline
		GEF \cite{GEF}   & 48.23 & 47.80 & 50.32 & 42.65 & 46.44 & 36.94 & 33.57 & 34.03 & 92.36 & 35.44 & 34.76 & 90.51 & 49.42 \\
		\hline
		\hline
		Our-I          & \textbf{60.44} & \textbf{54.92} & \textbf{54.78} & 46.04 & \textbf{53.90} & 44.59 & 34.82 & 42.28 & \textbf{93.63} & 36.69 & \textbf{45.30} & \textbf{95.25} & \textbf{55.22} \\
		\hline
		Our-II          & 59.39 & \textbf{57.97} & \textbf{56.05} & 45.41 & 52.54 & 47.77 & \textbf{35.62} & \textbf{45.09} & \textbf{95.54} & \textbf{38.20} & \textbf{45.30} & \textbf{95.93} & \textbf{56.24} \\
		\hline
	\end{tabular}
\end{table*}

\begin{table*}[!h]
	\caption{Average Classification Accuracy(\%) of Office-10 vs Caltech-10 with the DECAF-6 features}
	\label{tab2}
	\begin{tabular}{|c||c|c|c|c|c|c|c|c|c|c|c|c||c|}
		\hline
		Methods$/$Tasks & C$\rightarrow$A & C$\rightarrow$W & C$\rightarrow$D & A$\rightarrow$C & A$\rightarrow$W & A$\rightarrow$D & W$\rightarrow$C & W$\rightarrow$A & W$\rightarrow$D & D$\rightarrow$C & D$\rightarrow$A & D$\rightarrow$W & average \\
		\hline
		\hline
		ALEXNET \cite{AlexNet} & 91.90 & 83.70 & 87.10 & 83.00 & 79.50 & 87.40 & 73.00 & 83.80 & \textbf{100.0} & 79.00 & 87.10 & 97.70 & 86.10 \\
		\hline
		DDC \cite{DDC}    & 91.90 & 85.40 & 88.80 & 85.00 & 86.10 & 89.00 & 78.00 & 84.90 & \textbf{100.0} & 81.10 & 89.50 & 98.20 & 88.20 \\
		\hline
		DAN \cite{DAN}    & 92.00 & 90.60 & 89.30 & 84.10 & 91.80 & 91.70 & 81.20 & 92.10 & \textbf{100.0} & 80.30 & 90.00 & 98.50 & 90.10 \\
		\hline
		\hline
		JDA \cite{JDA}    & 89.70 & 83.70 & 86.60 & 82.20 & 78.60 & 80.20 & 80.50 & 88.10 & 100.0 & 80.10 & 89.40 & 98.90 & 86.50 \\
		\hline
		MEDA \cite{MEDA}   & 93.40 & 95.60 & 91.10 & 87.40 & 88.10 & 88.10 & \textbf{93.20} & \textbf{99.40} & 99.40 & 87.50 & 93.20 & 97.60 & \underline{92.80} \\
		\hline
		ARTL \cite{ARTL}   & 92.40 & 87.80 & 86.60 & 87.40 & 88.50 & 85.40 & 88.20 & 92.30 & \textbf{100.0} & 87.30 & 92.70 & \textbf{100.0} & 90.70 \\
		\hline
		VDA  \cite{VDA}   & 92.17 & 82.71 & 87.26 & 86.20 & 80.68 & 81.53 & 87.80 & 91.75 & \textbf{100.0} & 88.60 & 92.90 & 99.66 & 89.27 \\
		\hline
		SCA  \cite{SCA}   & 89.46 & 85.42 & 87.90 & 78.81 & 75.93 & 85.35 & 74.80 & 86.12 & \textbf{100.0} & 78.09 & 89.98 & 98.64 & 85.88 \\
		\hline
		JGSA \cite{JGSA}   & 91.44 & 86.78 & 93.63 & 84.86 & 81.02 & 88.54 & 84.95 & 90.71 & \textbf{100.0} & 86.20 & 91.96 & 99.66 & 89.98 \\
		\hline
		DICD \cite{DICD}   & 91.02 & 92.20 & 93.63 & 86.02 & 81.36 & 83.44 & 83.97 & 89.67 & \textbf{100.0} & 86.11 & 92.17 & 98.98 & 89.88 \\
		\hline
		GEF  \cite{GEF}   & 91.34 & 88.81 & 91.08 & 83.97 & 78.64 & 85.99 & 83.88 & 89.25 & \textbf{100.0} & 86.29 & 92.28 & 98.98 & 89.21 \\
		\hline
		\hline
		Our-I & \textbf{93.42} & \textbf{95.93} & \textbf{95.54} & \textbf{87.44} & \textbf{92.20} & \textbf{91.72} & 87.18 & 91.75 & \textbf{100.0} & 87.27 & \textbf{93.53} & \textbf{100.0} & \textbf{93.00} \\
		\hline
		Our-II & \textbf{93.42} & \textbf{95.93} & \textbf{96.82} & \textbf{88.42} & \textbf{92.88} & \textbf{91.72} & 88.87 & 92.17 & \textbf{100.0} & \textbf{88.87} & \textbf{93.63} & \textbf{100.0} & \textbf{93.56} \\
		\hline
	\end{tabular}
\end{table*}

\begin{table*}[!h]
	\caption{Average Classification Accuracy(\%) of Image-CLEF-DA and Office-31 with the RESNET-50 features}
	\label{tab3}
	\begin{tabular}{|c||c|c|c|c|c|c|c|c|c|c|c|c||c|}
		\hline
		Tasks$/$Methods & I$\rightarrow$P & P$\rightarrow$I & I$\rightarrow$C & C$\rightarrow$I & C$\rightarrow$P & P$\rightarrow$C & A$\rightarrow$D & A$\rightarrow$W & D$\rightarrow$A & D$\rightarrow$W & W$\rightarrow$D & W$\rightarrow$A & average \\
		\hline
		\hline
		MEDA \cite{MEDA}     & \textbf{80.20} & 91.50 & 96.20 & 92.70 & 79.10 & \textbf{95.80} & 85.30 & 86.20 & 72.40 & 97.20 & 99.40 & 74.00 & \underline{87.50} \\
		\hline
		RESNET-50 \cite{ResNet} & 74.80 & 83.90 & 91.50 & 78.00 & 65.50 & 91.20 & 68.90 & 68.40 & 62.50 & 96.70 & 99.30 & 60.70 & 78.45 \\
		\hline
		DAN  \cite{DAN}     & 74.50 & 82.20 & 92.80 & 86.30 & 69.20 & 89.80 & 78.60 & 80.50 & 63.60 & 97.10 & 99.60 & 62.80 & 81.42 \\
		\hline
		DANN \cite{DANN}     & 75.00 & 86.00 & 96.20 & 87.00 & 74.30 & 91.50 & 79.70 & 82.00 & 68.20 & 96.90 & 99.10 & 67.40 & 83.61 \\
		\hline
		RTN  \cite{RTN}     & 75.60 & 86.80 & 95.30 & 86.90 & 72.70 & 92.20 & 77.50 & 84.50 & 66.20 & 96.80 & 99.40 & 64.80 & 83.23 \\
		\hline
		JAN  \cite{JAN}     & 76.80 & 88.00 & 94.70 & 89.50 & 74.20 & 91.70 & 84.70 & 85.40 & 68.60 & 97.40 & 99.80 & 70.00 & 85.07 \\
		\hline
		CDAN \cite{CDAN}     & 76.70 & 90.60 & \textbf{97.00} & 90.50 & 74.50 & 93.50 & 89.80 & \textbf{93.10} & 70.10 & 98.20 & \textbf{100.0} & 68.00 & 86.83 \\
		\hline                   
		CAN  \cite{CAN}     & 78.20 & 87.50 & 94.20 & 89.50 & 75.80 & 89.20 & 85.50 & 81.50 & 65.90 & 98.20 & 99.70 & 63.40 & 84.05 \\
		\hline                   
		MADA \cite{MADA}     & 75.00 & 87.90 & 96.00 & 88.80 & 75.20 & 92.20 & 87.80 & 90.10 & 70.30 & 97.40 & 99.60 & 66.40 & 85.56 \\
		\hline
		\hline                   
		Our-I   & 79.50 & \textbf{92.00} & 95.67 & \textbf{93.17} & 78.33 & 95.50 & \textbf{90.36} & 88.43 & \textbf{74.09} & \textbf{98.74} & 99.80 & \textbf{74.76} & \textbf{88.36} \\ 
		\hline                  
		Our-II   & 79.33 & 88.83 & 95.67 & \textbf{93.17} & \textbf{79.33} & 91.83 & \textbf{90.76} & 88.93 & \textbf{75.43} & \textbf{98.49} & 99.80 & \textbf{75.15} & \textbf{88.06} \\
		\hline
	\end{tabular}
\end{table*}

\begin{table*}[!h]
	\caption{Average Classification Accuracy(\%) of Office-Home with the ResNet-50 features}
	\label{tab4}
	\begin{tabular}{|c||c|c|c|c|c|c|c|c|c|c|c|c||c|}
		\hline
		Tasks$/$Methods & A$\rightarrow$C & A$\rightarrow$P & A$\rightarrow$R & C$\rightarrow$A & C$\rightarrow$P & C$\rightarrow$R & P$\rightarrow$A & P$\rightarrow$C & P$\rightarrow$R & R$\rightarrow$A & R$\rightarrow$C & R$\rightarrow$P & average \\
		\hline
		\hline
		MEDA \cite{MEDA}     &  55.20 & 76.20 & 77.30 & 58.00 & \textbf{73.70} & 71.90 & 59.30 & 52.40 & 77.90 & 68.20 & 57.50 & 81.80 & 67.45 \\
		\hline
		RESNET-50 \cite{ResNet} &  34.90 & 50.00 & 58.00 & 37.40 & 41.90 & 46.20 & 38.50 & 31.20 & 60.40 & 53.90 & 41.20 & 59.90 & 46.13 \\
		\hline
		DAN \cite{DAN}      &  43.60 & 57.00 & 67.90 & 45.80 & 56.50 & 60.40 & 44.00 & 43.60 & 67.70 & 63.10 & 51.50 & 74.30 & 56.28 \\
		\hline
		DANN  \cite{DANN}    &  45.60 & 59.30 & 70.10 & 47.00 & 58.50 & 60.90 & 46.10 & 43.70 & 68.50 & 63.20 & 51.80 & 76.80 & 57.63 \\
		\hline
		JAN  \cite{JAN}     &  45.90 & 61.20 & 68.90 & 50.40 & 59.70 & 61.00 & 45.80 & 43.40 & 70.30 & 63.90 & 52.40 & 76.80 & 58.31 \\
		\hline
		CDAN \cite{CDAN}     &  50.70 & 70.60 & 76.00 & 57.60 & 70.00 & 70.00 & 57.40 & 50.90 & 77.30 & 70.90 & 56.70 & 81.60 & 65.81 \\
		\hline                                     
		MDD \cite{MDD}      &  54.90 & 73.70 & 77.80 & 60.00 & 71.40 & 71.80 & 61.20 & 53.60 & 78.10 & \textbf{72.50} & \textbf{60.20} & 82.30 & \underline{68.13} \\
		\hline                                     
		TADA \cite{TADA}     &  53.10 & 72.30 & 77.20 & 59.10 & 71.20 & 72.10 & 59.70 & 53.10 & 78.40 & 72.40 & 60.00 & 82.90 & 67.63 \\
		\hline                                     
		BSP  \cite{BSP}     &  52.00 & 68.60 & 76.10 & 58.00 & 70.30 & 70.20 & 58.60 & 50.20 & 77.60 & 72.20 & 59.30 & 81.90 & 66.25 \\
		\hline                                     
		TAT  \cite{TAT}     &  51.60 & 69.50 & 75.40 & 59.40 & 69.50 & 68.60 & 59.50 & 50.50 & 76.80 & 70.90 & 56.60 & 81.60 & 65.83 \\
		\hline
		\hline                   
		Our-I   & \textbf{58.44} & \textbf{77.79} & \textbf{79.32} & \textbf{61.60} & 72.81 & \textbf{73.03} & \textbf{62.71} & \textbf{55.33} & \textbf{78.91} & 70.42 & 60.09 & \textbf{83.24} & \textbf{69.47} \\
		\hline                  
		Our-II   & \textbf{57.18} & \textbf{76.86} & \textbf{78.93} & \textbf{61.23} & 72.36 & \textbf{72.60} & \textbf{62.30} & \textbf{54.20} & \textbf{79.37} & 70.58 & 60.11 & \textbf{83.17} & \textbf{69.07} \\
		\hline
	\end{tabular}
\end{table*}

\begin{table*}[!h]
	\caption{Average Classification Accuracy(\%) of Office-10 vs Caltech-10 with the SURF features}
	\label{tab5}
	\begin{tabular}{|c||c|c|c|c|c|c|c|c|c|c|c|c||c|}
		\hline
		Methods$/$Tasks & C$\rightarrow$A & C$\rightarrow$W & C$\rightarrow$D & A$\rightarrow$C & A$\rightarrow$W & A$\rightarrow$D & W$\rightarrow$C & W$\rightarrow$A & W$\rightarrow$D & D$\rightarrow$C & D$\rightarrow$A & D$\rightarrow$W & average \\
		\hline
		\hline
		D$_{tra}$   & 58.04 & 52.20 & 52.23 & \textbf{46.22} & 48.47 & 42.04 & \textbf{36.06} & 41.34 & 92.99 & 34.02 & 37.06 & 94.92 & 52.97 \\
		\hline
		Our-I          & \textbf{60.44} & \textbf{54.92} & \textbf{54.78} & 46.04 & \textbf{53.90} & \textbf{44.59} & 34.82 & \textbf{42.28} & \textbf{93.63} & \textbf{36.69} & \textbf{45.30} & \textbf{95.25} & \textbf{55.22} \\
		\hline
		\hline
		D$_{ter}$   & 53.55 & 55.25 & 54.78 & 42.65 & 49.83 & 45.86 & 34.82 & 42.17 & 92.36 & 34.91 & \textbf{46.14} & 93.22 & 53.79 \\
		\hline
		Our-II          & \textbf{59.39} & \textbf{57.97} & \textbf{56.05} & \textbf{45.41} & \textbf{52.54} & \textbf{47.77} & \textbf{35.62} & \textbf{45.09} & \textbf{95.54} & \textbf{38.20} & 45.30 & \textbf{95.93} & \textbf{56.24} \\
		\hline
		\hline
		D$_{tra}$+D$_{ter}$   & 57.20 & 52.20 & 50.96 & \textbf{46.13} & 48.47 & 42.04 & \textbf{35.98} & 41.44 & \textbf{92.99} & \textbf{34.82} & 38.31 & 94.92 & 52.95 \\
		\hline
	\end{tabular}
\end{table*}

\section{Experiments}

\subsection{Experimental Setup}
\label{Experimental Setup}
\par We adopted 4 popular image datasets concerning the cross-domain object recognition: 1) \textbf{Office-10 vs Caltech-10} \cite{GFK} consists of 10 common categories from 4 domains, i.e., Amazon (A), DSLR (D), Webcam (W), and Caltech (C), where the 800-dimension SURF features and the 4096-dimension DECAf-6 features were utilized; 2) \textbf{Image-CLEF-DA} \cite{CDAN} contains 12 shared classes pertaining to 3 domains, i.e., Caltech (C), ImageNet (I), and Pascal (P); 3) \textbf{Office-31} \cite{Office31} involves 31 common semantics belonging to 3 domains, i.e., Amazon (A), DSLR (D), and Webcam (W); 4) \textbf{Office-Home} \cite{DHN} includes 15500 images of 65 common classes from 4 domains, i.e., Art (A), Clipart (C), Product (P), and Real-world (R). Notably, the 2048-dimension RESNET-50 features were employed for the last 3 datasets. Within each dataset, any two domains could be the source and target domains to establish different DA tasks.

\par The proposed approach with two different strategies (i.e., Our-I, Our-II) were compared with several state-of-the-art DA methods: 1) the traditional DA methods: Joint Distribution Adaptation (JDA) \cite{JDA}, Balanced Distribution Adaptation (BDA) \cite{BDA}, Manifold Embedded Distribution Alignment (MEDA) \cite{MEDA}, Adaptation Regularization Transfer Learning (ARTL) \cite{ARTL}, Visual Domain Adaptation (VDA) \cite{VDA}, Scatter Component Analysis (SCA) \cite{SCA}, Joint Geometrical and Statistical Alignment (JGSA) \cite{JGSA}, Domain Invariant and Class Discriminative (DICD) \cite{DICD}, Transfer Independently Together (TIT) \cite{TIT}, Graph embedding Framework (GEF) \cite{GEF}; 2) the deep DA methods: ALEXNET \cite{AlexNet}, Deep Domain Confusion (DDC) \cite{DDC}, Deep Adaptation Networks (DAN) \cite{DAN}, RESNET-50 \cite{ResNet}, Domain Adversarial training of Neural Networks (DANN) \cite{DANN}, Residual Transfer Networks (RTN) \cite{RTN}, Joint Adaptation Networks (JAN) \cite{JAN}, Conditional Domain Adversarial Network (CDAN) \cite{CDAN}, Collaborative Adversarial Network (CAN) \cite{CAN}, Multi-Adversarial Domain Adaptation (MADA) \cite{MADA}, Margin Disparity Discrepancy (MDD) \cite{MDD}, Transferable Attention for Domain Adaptation (TADA) \cite{TADA}, Batch Spectral Penalization (BSP) \cite{BSP}, Transferable Adversarial Training (TAT) \cite{TAT}.


\par Parameters setting: we first fixed $T=5$, $p=20$ for all DA tasks, then we set $k=20, \alpha=0.05$ for Office-10 vs Caltech-10 and Image-CLEF-DA datasets, but $k=100, \alpha=0.1$ for Office-31 and Office-Home datasets since they contain more categories. For $\beta, \lambda$ introduced in this paper, we simply set them by searching the small discrete ranges and used the optimal ones, i.e., $\beta\in [-1.0, -0.9, ..., 0, 0.1, ... , 0.9, 1.0], \lambda\in [0.2, 0.3, ..., 1.0]$. 

\subsection{Experimental Results}

\par The classification accuracy results on the datasets of  Office-10 vs Caltech-10 dataset with SURF and DECAF-6 features, Image-CLEF-DA, Office-31, Office-Home are illustrated in Table. \ref{tab1} $\sim$ Table. \ref{tab4}, and it could be made several observations from these results. Firstly, the results of our approach with two different strategies, i.e., Our-I (Section \ref{first}), Our-II (Section \ref{second}), are both substantially higher than all other traditional and deep DA approaches on most DA tasks (36/48, 31/48 tasks), and the average results in terms of different datasets are 55.22\%/56.24\%, 93.00\%/93.56\%, 88.36\%/88.06\%, 69.47\%/69.07\%, which have 2.52\%/3.54\%, 0.20\%/0.76\%, 0.86\%/0.56\%, 1.34\%/0.94\% improvements compared with the best baseline methods (underlined ones). Secondly, the results on the Office-Home dataset verify that our method could be also applicable to the large-scale dataset, and be able to achieve favorable results. Moreover, the deep DA methods have to estimate very large amounts of hyperparameters, be it either automatic or manual, which is inherently time-consuming, laborious, and prone to errors. Differently, the proposed approach only involves several parameters that could easily be set by human experience or cross-validation, which further implies the efficiency and robustness of the proposed approach in cross-domain object recognition problems.  

\begin{figure*}[h]
	\centering
	\includegraphics[width=1.0\linewidth,height=0.3\textheight]{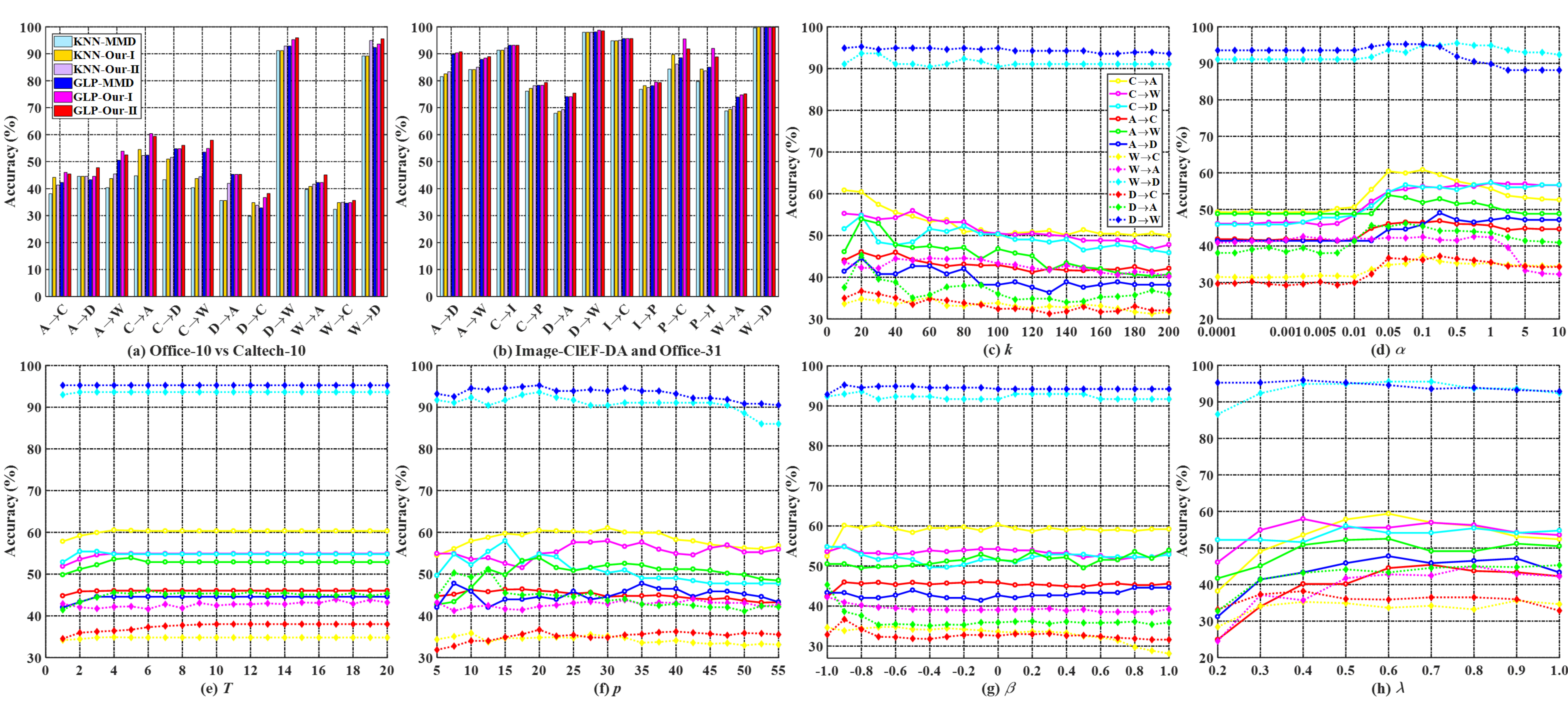}
	\caption{Accuracy results of the original MMD and ours using KNN and GLP classifiers on three different datasets (i.e., (a), (b)). Parameter sensitivity of $T$, $k$, $\alpha$, $p$, $\beta$ and $\lambda$ values (i.e., (c) $\sim$ (h)).}
	\label{fig2}
\end{figure*}

\begin{figure*}[h]
	\centering
	\includegraphics[width=1.0\linewidth,height=0.3\textheight]{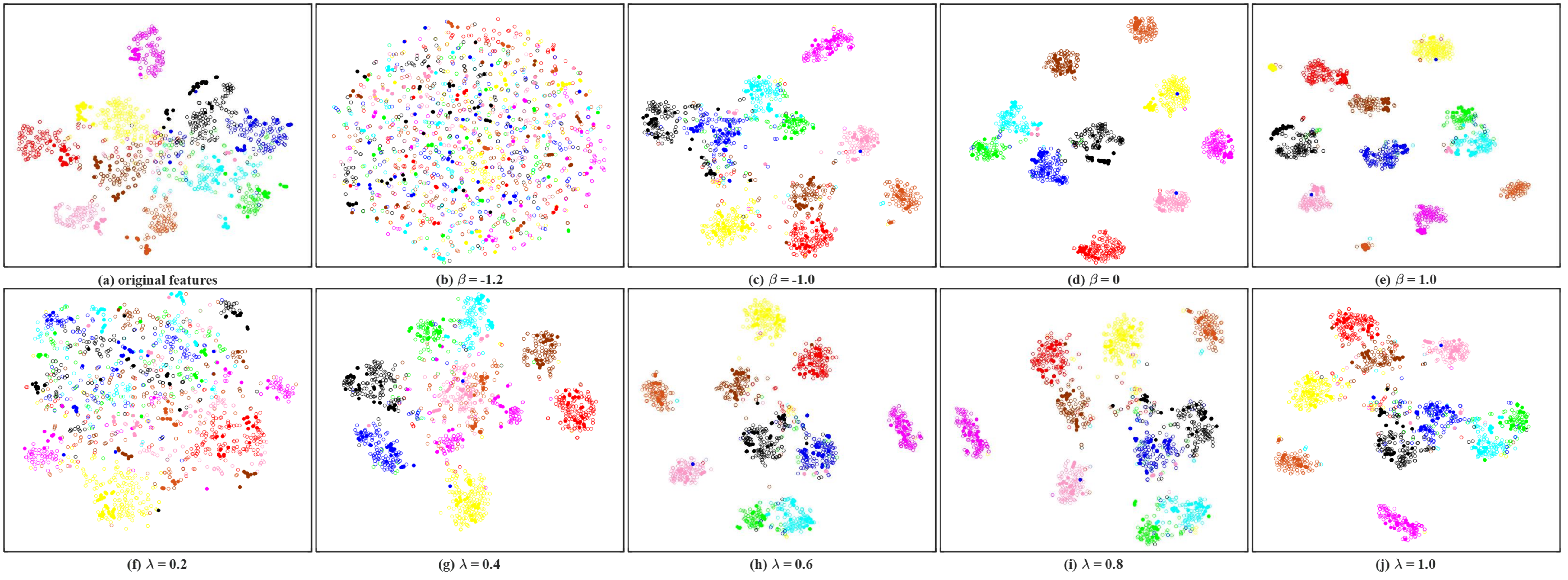}
	\caption{tSNE visualization on C$\rightarrow$W of the Office-10 vs Caltech-10 dataset with DECAF-6 features. The source domain is denoted by hollow circles and the target domain is denoted by solid circles, and different colors represent various categories.}
	\label{fig3}
\end{figure*}

\begin{figure}[h]
	\centering
	\includegraphics[width=1.0\linewidth,height=0.25\textheight]{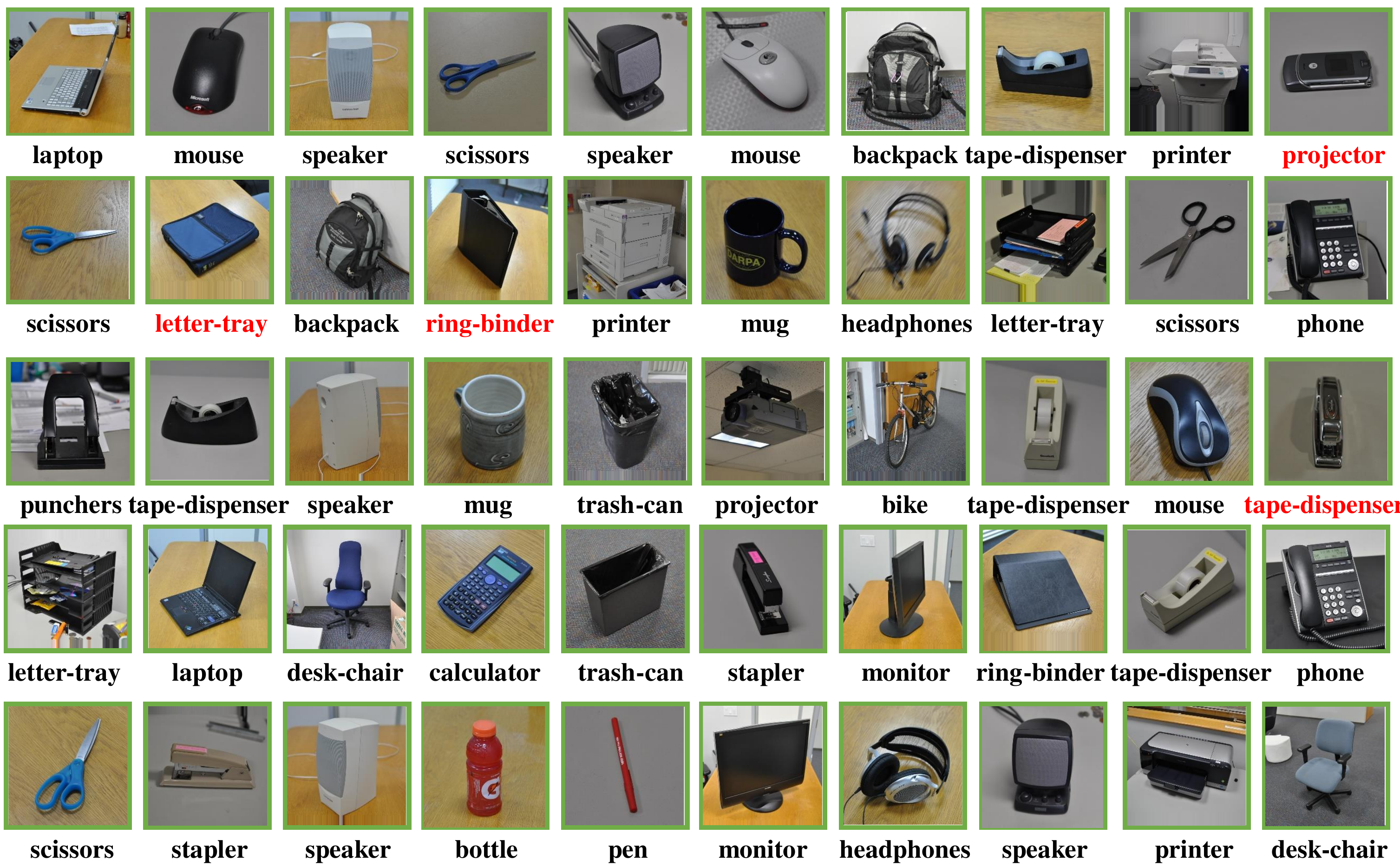}
	\caption{Recognition results of Our-I, correct and incorrect labeled instances are marked in black and red fonts.}
	\label{fig4}
\end{figure}

\begin{figure}[h]
	\centering
	\includegraphics[width=1.0\linewidth,height=0.25\textheight]{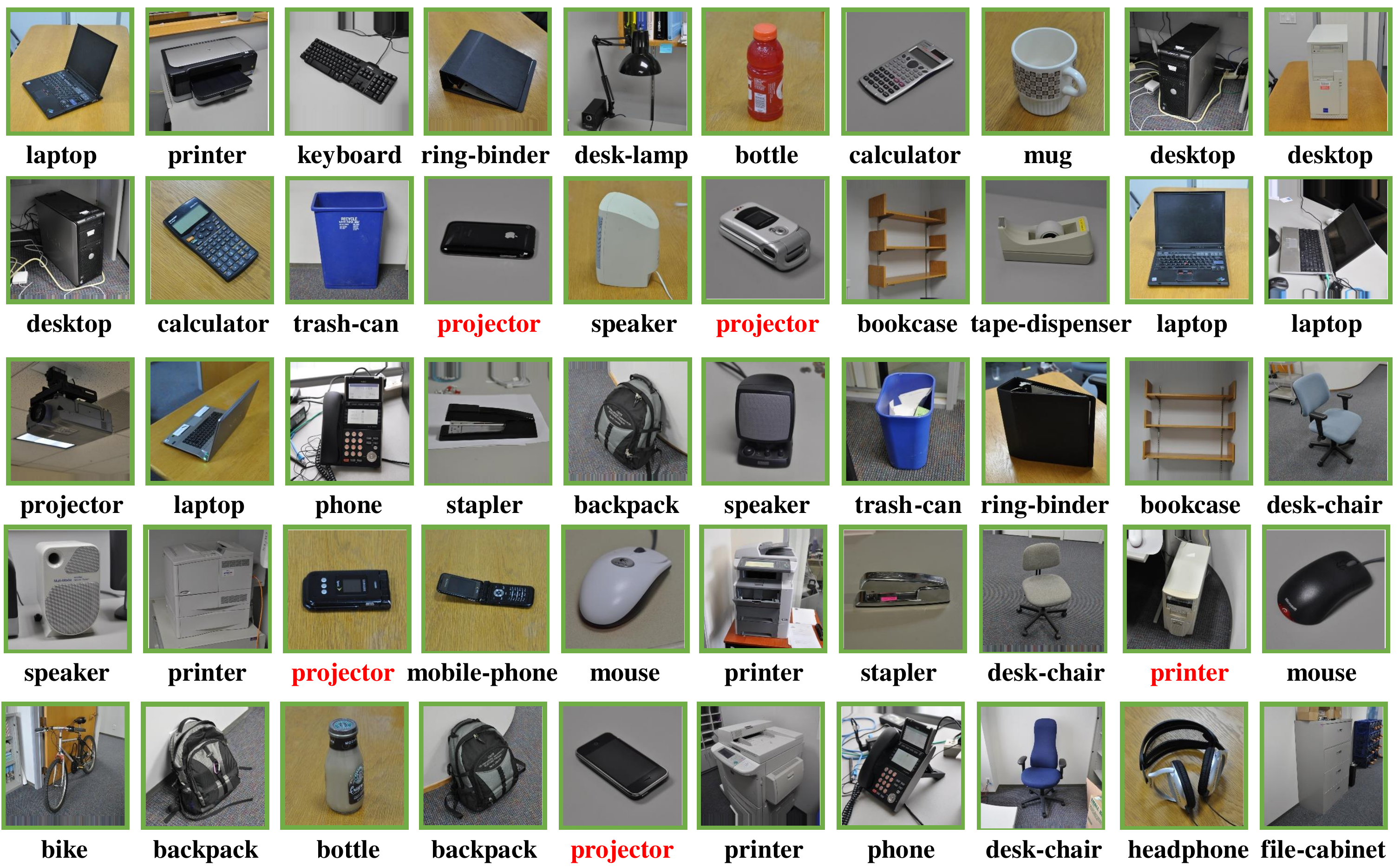}
	\caption{Recognition results of Our-II, correct and incorrect labeled instances are marked in black and red fonts.}
	\label{fig5}
\end{figure}

\subsection{Ablation Study} 
\label{Ablation Study}
\par To prove the effectiveness of our approach compared with the original MMD, we report their accuracy results using KNN (i.e., K-Nearest Neighbor, K=1) and GLP classifiers on the datasets of Office-10 vs Caltech-10, Image-CLEF-DA, and Office-31. As shown in Fig. \ref{fig2} (a), (b), the GLP (i.e., the darker color) usually performs better than KNN (i.e., the lighter color) on most DA tasks. Remarkably, our approach with two different strategies could achieve higher performances than the original MMD on nearly all evaluations regardless of the classifiers, thus it is capable of mitigating the adverse impacts of the original MMD on the feature discriminability.  

\par The discriminative DA approaches promoted the feature discriminability usually by accumulating to consider the intra-class and inter-class distances (i.e., $MMD+\gamma_1D_{tra}+\gamma_2D_{ter}$). However, the parameters involved in their models are often set in unknown regions and redundant. Specifically, $\gamma_1$ and $\gamma_2$ are empirically set as $0.01,0.01$. This paper theoretically proves that there exist special weights that are implicit in the MMD. Then, we randomly choose a group of implicit weights in the DA task of C$\rightarrow$A, i.e., [0.012, 0.020, 0.021, 0.015, 0.026, 0.019, 0.016, 0.027, 0.031, 0.026], and it could be observed that it is very consistent with their empirical values. This provides us the theoretical guidance for the optimal parameter regions, and intensifying the feature discriminability more correctly. Moreover, once the optimal parameter regions are revealed, it is easy to set the two parameters introduced in this paper, since their physical meanings are pretty clear. 


\par We denote the variants $MMD+0.01D_{tra}$, $MMD+0.01D_{ter}$, $MMD+0.01D_{tra}+0.01D_{ter}$ as $D_{tra}$, $D_{ter}$, $D_{tra}+D_{ter}$, respectively. As reported in Tab. \ref{tab5}, we observe that our approach could give better results over the corresponding discriminative DA methods. We speculate that it is because our approach could search for more correct weights with clear physical meanings. Furthermore, the results of $D_{tra}+D_{ter}$ slightly go down compared with $D_{tra}$ and $D_{ter}$, which could further verify that the relationship between them is as one falls, another rises mentioned before, thus this manner does not necessarily increase the DA performances.

\par Following the work in \cite{DAN, WMMD}, we visualize the features learned by the proposed approach with different settings on the task C$\rightarrow$W of the Office-10 vs Caltech-10 dataset with DECAf-6 features. As illustrated in Fig. \ref{fig3} (a), the original features perform badly in both feature transferability and its discriminability. Fig. \ref{fig3} (b) excessively maximizes the intra-class distance in the MMD so that different classes are very complicated and confused. Fig. \ref{fig3} (c) is the original MMD, and different classes still overlap with each other, which produces an unexpected degradation of the feature discriminability. Fig. \ref{fig3} (d) precisely removes the term of intra-class distance in the MMD so that the negative impacts on the feature discriminability is mitigated significantly. Fig. \ref{fig3} (e) overly considers the feature discriminability, thus the cross-domain distributional alignment is slightly weakened. The last 5 cases in Fig. \ref{fig3} are corresponding to varying values of $\lambda$ in the proposed second strategy, and the larger its value is, the more significant feature transferability is. Fig. \ref{fig3} (f) behaviors worst among them since its transferability is overly undermined. With the increase of $\lambda$, Fig. \ref{fig3} (h) achieves the best performance but it begins to slide in Fig. \ref{fig3} (i) and Fig. \ref{fig3} (j) since they heavily concentrate the transferability, which can verify that our approach could effectively leverage the importance of feature transferability and its discriminability. Generally speaking, these observations could prove the theoretical results in this paper.

\par We conducted the experiments of parameters sensitivity, with varying values of one parameter after fixing the others discussed in Section \ref{Experimental Setup}, on 12 DA tasks constructed from the Office-10 vs Caltach-10 dataset with the SURF features, while similar trends on all other cross-domain datasets are not shown due to space limitations. As illustrated in Fig. \ref{fig2} (c) $\sim$ Fig. \ref{fig2} (h), we plot classification results w.r.t., their different values, and choose $T\in [1,20]$, $k\in [10,200]$, $\alpha\in [0.0001,10]$, $p\in [5,55]$, $\beta\in [-1.0,1.0]$, $\lambda\in [0.2,1.0]$, and the fisrt 5 parameters are only evaluated by Our-I. 

\par Specifically, Fig. \ref{fig2} (a) indicates that the accuracy results achieve better performances steadily within only 5 iterations. For $k$ and $\alpha$, the scale of projection $\textbf{\textit{A}}$ increases with $k$ enlarges, and $\alpha$ aims to control its scale. Therefore, their specific accuracy curves have a similar trend of the first rise up and then move down when we fix one of them (i.e., Fig. \ref{fig2} (b) and Fig. \ref{fig2} (c)). Fig. \ref{fig2} (d) shows that the optimal results could be achieved under a wide range of $p$. 

\par Concerning $\beta$ and $\lambda$, they measure the relative importance between the feature transferability and its discriminability. Notably, the significance of feature discriminability is augmented with $\beta$ increases but $\lambda$ decreases, and the original MMD is the extreme case when $\beta=-1.0$ or $\lambda=1.0$. Fig. \ref{fig2} (e) and Fig. \ref{fig2} (f) illustrates that those two essential properties have different relative importance on the different DA tasks, and the two curves both have the feature of the first rise up and then move down. These observations can verify that the proposed novel insight of the MMD could prepare us to study their relative importance conveniently.

\subsection{Recognition Results} 
\par We conducted the proposed approach (i.e., Our-I, Our-II) on the task A$\rightarrow$D from Office-31 dataset, and randomly select 50 target images with their predictive results. As shown in Fig. \ref{fig4} and Fig. \ref{fig5}, our method could obtain desirable results where the correct and incorrect labeled instances are marked in black and red fonts.

\section{Conclusion}
\par In this paper, we theoretically prove the working principle of the
MMD that is highly consistent with the transferability of human
beings, which also illustrates the reasons for degradation of the feature discriminability in the MMD, and provides qualitative
and quantitative guidance in studying the relationship between
the feature transferability and its discriminability. Based on this, we propose a novel discriminative MMD with two different strategies, where we consider the intra-class and inter-class distances alone since we prove that their relationship is as one falls, another rises, thus the redundant parameter could be further omitted. The experiments on several benchmark datasets prove the validity of theoretical results and demonstrate that the proposed approach could perform better than the comparative state-of-art DA methods (traditional and deep DA approaches) markedly.   


%

%

\section*{Acknowledgment}

This work was supported in part by the National Natural Science Foundation of China (NSFC) under Grants No.61772108, No.61932020, and No.61976038, No.U1908210.

\ifCLASSOPTIONcaptionsoff
  \newpage
\fi



%
%
%
\bibliographystyle{IEEEtran}
\bibliography{bare_jrnl.bib}

%

\begin{IEEEbiography}[{\includegraphics[width=1in,height=1.25in,clip,keepaspectratio]{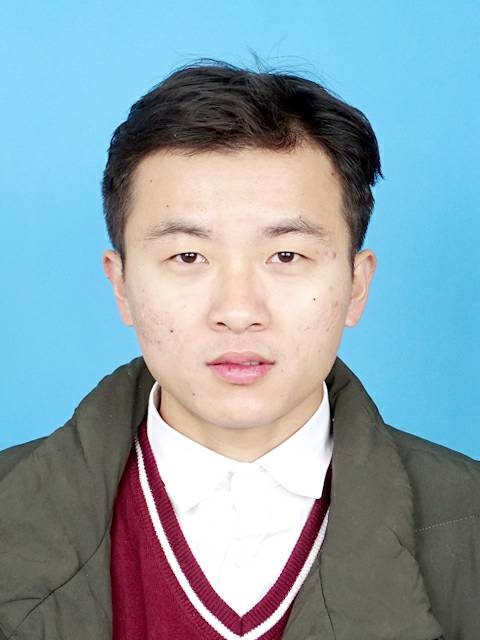}}]{Wei Wang}
is currently a Ph.D. candidate at the School of Software Technology, Dalian University of Technology, Dalian, China. He received the B.S. degree at the school of science from the Anhui Agricultural University, Hefei, China, in 2015. He received the M.S. degree at the school of computer science and technology from the Anhui University, Hefei, China, in 2018. His major research interests include  transfer learning, zero-shot learning, deep learning, etc.
\end{IEEEbiography}
\begin{IEEEbiography}[{\includegraphics[width=1in,height=1.25in,clip,keepaspectratio]{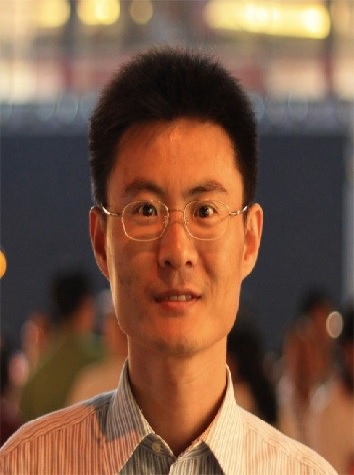}}]{Haojie Li}
	is currently a Professor in the DUT-RU International School of Information Science \& Engineering, Dalian University of Technology. He received the B.E. and the Ph.D. degrees from Nankai University, Tianjin and the Institute of Computing Technology, Chinese Academy of Sciences, Beijing, in 1996 and 2007 respectively. From 2007 to 2009, he was a Research Fellow in the School of Computing, National University of Singapore. He is a member of IEEE and ACM. His research interests include social media computing and multimedia information retrieval. He has co-authored over 70 journal and conference papers in these areas, including IEEE TCSVT, TMM, TIP, ACM Multimedia, ACM ICMR, etc.
\end{IEEEbiography}
\begin{IEEEbiography}[{\includegraphics[width=1in,height=1.25in,clip,keepaspectratio]{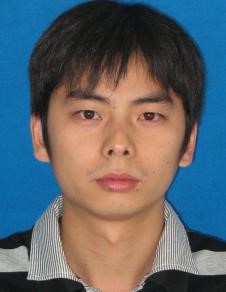}}]{Zhengming Ding}
	received the B.Eng. degree in information security and the M.Eng. degree in computer software and theory from University of Electronic Science and Technology of China (UESTC), China, in 2010 and 2013, respectively. He received the Ph.D. degree from the Department of Electrical and Computer Engineering, Northeastern University, USA in 2018. He is a faculty member affiliated with Department of Computer, Information and Technology, Indiana University-Purdue University Indianapolis since 2018. His research interests include transfer learning, multi-view learning and deep learning. He received the National Institute of Justice Fellowship during 2016-2018. He was the recipients of the best paper award (SPIE 2016) and best paper candidate (ACM MM 2017). He is currently an Associate Editor of the Journal of Electronic Imaging (JEI). He is a member of IEEE.
\end{IEEEbiography}
\begin{IEEEbiography}[{\includegraphics[width=1in,height=1.25in,clip,keepaspectratio]{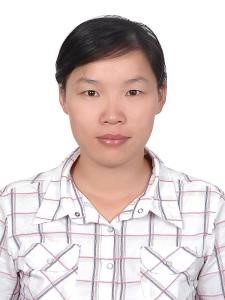}}]{Zhihui Wang}
	received the B.S. degree in software engineering in 2004 from the North Eastern University, Shenyang, China. She received her M.S. degree in software engineering in 2007 and the Ph.D degree in software and theory of computer in 2010, both from the Dalian University of Technology, Dalian, China. Since November 2011, she has been a visiting scholar of University of Washington. Her current research interests include image processing and image compression.
\end{IEEEbiography}






\end{document}